\documentclass[letterpaper]{article}
\usepackage{aaai}
\usepackage{times}
\usepackage{helvet}
\usepackage{courier}
\usepackage{graphicx}
\usepackage{multirow}
\begin{document}
%
\title{Style-transfer and Paraphrase: Looking for a Sensible Semantic Similarity Metric}
\author{}
\author{Ivan P. Yamshchikov\\
Max Planck Institute for \\
Mathematics in the Sciences,\\
Leipzig, Germany\\
\texttt{ivan@yamshchikov.info}\And
Viacheslav Shibaev\\
Ural Federal University\\
Ekaterinburg, Russia\And
Nikolay Khlebnikov\\
Ural Federal University\\
Ekaterinburg, Russia\And
Alexey Tikhonov\\
Yandex,\\
Berlin, Germany\\
}

\maketitle
\begin{abstract}
\begin{quote}
The rapid development of such natural language processing tasks as style transfer, paraphrase, and machine translation often calls for the use of semantic similarity metrics. In recent years a lot of methods to measure the semantic similarity of two short texts were developed. This paper provides a comprehensive analysis for more than a dozen of such methods. Using a new dataset of fourteen thousand sentence pairs human-labeled according to their semantic similarity, we demonstrate that none of the metrics widely used in the literature is close enough to human judgment in these tasks. A number of recently proposed metrics provide comparable results, yet Word Mover Distance is shown to be the most reasonable solution to measure semantic similarity in reformulated texts at the moment. 
\end{quote}
\end{abstract}

\section{Introduction}

Style transfer and paraphrase are two tasks in Natural Language Processing (NLP). Both of them are centered around the problem of an automated reformulation. Given an input text, the system tries to produce a new rewrite that resembles the old text semantically. In the task of paraphrase, semantic similarity is the only parameter that one tries to control. Style transfer usually controls more aspects of the text and could, therefore, be regarded as an extension of a paraphrase. Intuitive understanding of style transfer problem is as follows: if an input text has some attribute $A$, say, politeness, a system generates new text similar to the input semantically but with attribute $A$ changed to the target $\tilde{A}$.  For example, given a polite sentence "could you be so kind, give me a hand" and a target "not polite" the system produces a rewrite "God damn, help me".   

The significant part of current works perform style transfer via an encoder-decoder architecture with one or multiple style discriminators to learn disentangled representations \cite{hylsx}. This basic architecture can have various extensions, for example, can control POS-distance between input and output \cite{tian18}, or have additional discriminator or an extra loss term to improve the quality of the latent representations \cite{yamshchikov2019decomposing}. There are also other approaches to this problem that do not use ideas of disentangled latent representations but rather treat it as a machine translation problem; see, for example, \cite{subramanian18}. However, independently of a chosen architecture, one has to control the semantic component of the output text. It is expected to stay the same as the system changes the style of the input. This aspect makes the problem of style transfer naturally related to the problem of paraphrase \cite{para1}, \cite{para2}, \cite{para3}. It also raises the question of how one could automatically measure the semantic similarity of two texts in these problems.

As with every NLP task that is relatively new, the widely accepted baselines and evaluations metrics are still only emerging. There are ongoing discussions on which aspects of the texts are stylistic and could be changed by the style transfer system and which are semantic and therefore are technically out of the scope of the style transfer research \cite{TYwrong}. This paper refrains from these discussions. It instead attempts to systematize existing methods of quality assessment for the tasks of style transfer that are used in different state of art research results. We also put these methods into the perspective of paraphrase tasks. To our knowledge, that was not done before. The contribution of the paper is four-fold: 

\begin{itemize}
\item it compares more than a dozen of existing semantic similarity metrics used by different researchers to measure the performance of different style transfer methods; 
\item using human assessment of 14 thousand pairs of sentences it demonstrates that there is still no optimal semantic-preservation metric that could be comparable with human judgment in context of paraphrase and textual style transfer, however Word Mover Distance \cite{kusner2015word} seems to be the most promising one;
\item it proposes a simple necessary condition that a metric should comply with to be a valid semantic similarity metric for the task of style transfer; 
\item it shows that some metrics used in style transfer literature should not be used in the context of style transfer at all.
\end{itemize} 

\section{Measuring semantic preservation}
\label{Sec:mes}

Style transfer, as well as a paraphrase, naturally demands the preservation of the semantic component as the input sentence is transformed into the desired output. Different researchers use different methods to measure this preservation of semantics. 

Despite its disadvantages \cite{larsson2017disentangled}, one of the most widely used semantic similarity metrics is BLEU. \cite{tikhonov2019style} show that it could be manipulated in a way that the system would show higher values of BLEU on average, producing sentences that are completely detached from the input semantically. However, BLEU is easy to calculate and is broadly accepted for various NLP tasks that demand semantic preservation \cite{vaswani2017attention}, \cite{hylsx}, \cite{cohn2019lost}. Alongside BLEU, there are other, less broadly accepted metrics for semantic preservation. For example, \cite{zhang2018shaped} work with different versions of ROUGE. 

\cite{fu2}, \cite{john18} or \cite{romanov18} compute a sentence embedding by concatenating the min, max, and mean of its word embeddings and use the cosine similarity between the source and generated sentence embeddings as an indicator of content preservation. \cite{tian18} uses POS-distance alongside with BLEU and BLEU between human-written reformulations and the actual output of the system. 

One of the most recent contributions in this area \cite{mir2019evaluating} evaluates several of the metrics mentioned above as well as METEOR \cite{banerjee2005meteor} and Word Mover's Distance (WMD). This metric is calculated as the minimum "distance" between word embeddings of input and output \cite{kusner2015word}. 

In this paper, we use these metrics of content-preservation listed above alongside with several others that are used for semantic similarity in other NLP tasks recently. We put all these metrics into the context of paraphrase and style transfer. These metrics are:
\begin{itemize}
 \item POS-distance that looks for nouns in the input and output and is calculated as a pairwise distance between the embeddings of the found nouns;
 \item Word overlap calculated as a number of words that occur in both texts;
 \item chrF \cite{popovic2015chrf} -- a character n-gram F-score that measures number of n-grams that coincide in input and output;
 \item cosine similarity calculated in line with \cite{fu2} with pre-trained embeddings by GloVe \cite{pennington2014glove};
 \item cosine similarity calculated similarly but using FastText word embeddings \cite{joulin2016fasttext};
 \item L2 distance based on ELMo \cite{peters2018deep}
 \item WMD \cite{kusner2015word} that defines the distance between two documents as an optimal transport problem between the embedded words;
 \item BLEU \cite{papineni2002bleu};
 \item ROUGE-1 \cite{lin2000automated} compares any text to any other (typically human-generated) summary using a recall-oriented approach and unigrams;
 \item ROUGE-2 that uses bigrams;
 \item ROUGE-L \cite{lin2004automatic} that identifies longest co-occurring in sequence n-grams;
 \item Meteor \cite{banerjee2005meteor} metric that is based on a harmonic mean of unigram precision and recall, with recall weighted higher than precision and some additional features, such as stemming and synonymy matching;
 \item and the BERT score proposed in \cite{zhang2019bertscore} for the estimation of the generated texts.
\end{itemize}

All these metrics are known to vary from dataset to dataset but show consistent results within one data collection. In the next section, we try to come up with a set of various paraphrases and style transfer datasets that would allow us to see qualitative differences between these metrics of semantic similarity.

\section{Data}

The task of paraphrasing a given sentence is better formalized than the task of style transfer. However, to our knowledge, there were no attempts to look at these two tasks in one context. There are several datasets designed to benchmark semantic similarity metrics. The most widely used is STS-B, see \cite{cer2017semeval}. \cite{zhang2019paws} provide a dataset of sentences that have high lexical overlap without being paraphrases. Quora Questions Paraphrase dataset\footnote{https://www.kaggle.com/quora/question-pairs-dataset} provides paraphrases of Quora questions. However, these datasets do not include style transfer examples whereas the focus of this paper is to align semantic similarity metrics used for paraphrase with the one used in style transfer community. Here we intend to work with the metrics listed in the previous section and calculate them over three paraphrase and two style transfer datasets that are often used for these two NLP tasks. The paraphrase datasets include: 

\begin{itemize}
\item different versions of English Bibles \cite{carlson2017zero};
\item English Paralex dataset\footnote{http://knowitall.cs.washington.edu/paralex/};
\item English Paraphrase dataset\footnote{http://paraphrase.org}.
\end{itemize}

The style transfer datasets are:
\begin{itemize}
\item Dataset of politeness introduced in \cite{rao2018dear} that we in line with the original naming given by the authors refer to as GYAFC later on;
\item Yelp! Reviews\footnote{https://www.yelp.com/dataset} enhanced with human written reviews with opposite sentiment provided by \cite{tian18}.
\end{itemize}

We suggest to work with these datasets, since they are frequently used for baseline measurements in paraphrase and style transfer literature.

Out of all these listed datasets we sample 1000 sentence pairs, where each pair of sentences consists of two paraphrases or two sentences with different style and comparable semantics. Experimental results that follow present averages of every measure of semantic similarity over these 1000 pairs for every dataset. Additionally to the paraphrases and style-transfer datasets we provide several datasets that consist of sentence pairs that have no common semantic component yet are sampled from the same datasets. We do it for several reasons: first, semantic similarity measure should be at least capable to distinguish sentence pairs that have no semantic similarity whatsoever from paraphrases or style-transfer examples, second, variation of the semantic similarity on random pairs for various corpora could show how a given metric depends on the corpus' vocabulary. These random datasets could be used as a form of a benchmark to estimate 'zero' for every semantic similarity metric.

All the metrics that we include in this paper already have undergone validation. These metrics hardly depend on the size of the random data sample provided it is large enough. They are also known to vary from one dataset to another. However, due to the laborious nature of this project, we do not know of any attempts to characterize these differences across various datasets. 

\section{Assessment}

This paper is focused on the applications of semantic similarity to the tasks of style transfer and paraphrase, however there are more NLP tasks that depend on semantic similarity measures. We believe that the reasoning and measurements presented in this paper are general enough to be transferred to other NLP tasks that depend upon a semantic similarity metric.

Table \ref{tab:ex1} and Table \ref{tab:ex2} show the results for fourteen datasets and thirteen metrics as well as the results of the human evaluation of semantic similarity. It is essential to mention that \cite{rao2018dear} provide different reformulations of the same text both in an informal and formal style. That allows us to use the GYAFC dataset not only as a style transfer dataset but also as a paraphrase dataset, and, therefore, extend the number of datasets in the experiment. To stimulate further research of semantic similarity measurements, we publish\footnote{https://github.com/VAShibaev/semantic\_similarity\_metrics}
our dataset that consists of 14 000 different pairs of sentences alongside with semantic similarity scores given by the annotators. Each sentence was annotated by at least three humans independently. There were 300+ English native speakers involved in the assessment. Every annotator was presented with two parallel sentences and was asked to assess how similar their meaning is. We used AmazonTurk with several restrictions on the turkers: these should be native speakers of English in the top quintile of the internal rating. Humans were to assess "how similar is the meaning of these two sentences" on a scale from $1$ to $5$. This is a standard formulation of semantic-similarity assessment task on AmazonTurk. Since annotators with high performance scores already know this task, we didn't change this standard formulation to ensure that gathered data is representative for standard semantic similarity problems. We publish all scores that were provided by the annotators to enable further methodological research. We hope that this dataset could be further used for a deeper understanding of semantic similarity.

\begin{table*}[t!]
\centering
\tiny{\begin{tabular}{lllllllll}
 Dataset & Human Labeling & POS-distance &	Word & chrF & Cosine	& Cosine &	WMD \\
  & & & overlap & & Similarity	& Similarity & \\
 & & & && Word2Vec	& FastText & \\
 \hline
Bibles & $3.54 \pm 0.72$ & $2.39 \pm 3.55$	& $0.47 \pm 0.18 $ & $0.54 \pm 0.18$	& $0.04 \pm 0.04$ &	$0.04\pm 0.02$ &	$0.57 \pm 0.29$ \\
Paralex & $3.28 \pm 0.8$ & $2.91 \pm 4.28$ & $0.43 \pm 0.18$ & 	$0.48 \pm 0.18$	& 	$0.13 \pm 0.09$ & 	$0.09 \pm 0.04$ & 	$0.62 \pm 0.3$ \\
Paraphrase & $3.6 \pm 0.79$ & $2.29 \pm 2.85$ & $0.31 \pm 0.2$ 	& $0.41 \pm 0.23$ & $0.29\pm 0.17$ 	& $0.21\pm 0.12$ \ & $0.77 \pm 0.34$\\
GYAFC formal & $3.63 \pm 0.75$ & $2.27 \pm 3.97$ & $0.5 \pm 0.22$	& $0.53 \pm 0.22$ & $0.06\pm 0.04$ 	& $0.05\pm 0.03$ \ & $0.57 \pm 0.35$\\
GYAFC informal & $3.41 \pm 0.78$ & $3.79 \pm 4.54$ & $0.32 \pm 0.17$ & $0.34 \pm 0.17$ & $0.09\pm 0.05$ 	& $0.09\pm 0.04$ \ & $0.76 \pm 0.31$\\
 \hline
Yelp! rewrite & $2.68 \pm 0.83$ & $1.11 \pm 2.34$ & $0.45 \pm 0.25$ &	$0.51 \pm 0.23$ & $0.08	\pm 0.06$ & $0.08 \pm 0.06$ & $0.61 \pm 0.31$ \\
GYAFC rewrite & $3.83 \pm 0.75$ & $2.32 \pm 3.91$ & $0.47 \pm 0.21$ & $0.53 \pm 0.22$ & $0.06\pm 0.04$ 	& $0.06\pm 0.04$ \ & $0.54 \pm 0.35$\\
\hline
Bibles random & $2.32 \pm 0.69$ & $11.21 \pm 8.08$ & $0.10 \pm 0.04$	& $0.17 \pm 0.05$ & $0.10\pm 0.07$ 	& $0.1\pm 0.03$ \ & $1.23 \pm 0.07$\\

Paralex random & $1.95 \pm 0.71$ & $10.31 \pm 4.84$ & $0.13 \pm 0.08$	& $0.14 \pm 0.05$ & $0.24\pm 0.09$ 	& $0.18\pm 0.04$ \ & $1.3 \pm 0.07$\\

Paraphrase random & $1.97 \pm 0.65$ & $7.47 \pm 2.5$ & $0.02 \pm 0.06$	& $0.1 \pm 0.05$ & $0.58\pm 0.18$ 	& $0.46\pm 0.13$ \ & $0.34 \pm 0.07$\\

GYAFC random & $2.13 \pm 0.73$ & $10.61 \pm 7.2$ & $0.05 \pm 0.05$	& $0.13 \pm 0.04$ & $0.15\pm 0.05$ 	& $0.15\pm 0.04$ \ & $1.24 \pm 0.08$\\
informal & & & & 	& & \\
GYAFC random & $2.12 \pm 0.74$ & $10.82 \pm 8.64$ & $0.08 \pm 0.05$ & $0.14 \pm 0.04$ & $0.15\pm 0.04$ 	& $0.14\pm 0.03$ \ & $1.26 \pm 0.07$\\
formal & & & & 	& & \\
GYAFC random & $2.07 \pm 0.7$ & $10.58 \pm 8.03$ & $0.06 \pm 0.05$	& $0.13 \pm 0.04$ & $0.15\pm 0.04$ 	& $0.14\pm 0.03$ \ & $1.25 \pm 0.07$\\
rewrite & &	& & 	& & \\
Yelp! random & $2.14 \pm 0.79$ & $8.97 \pm 4.35$ & $0.06 \pm 0.06$ & $0.14 \pm 0.04$ & $0.19 \pm 0.06$	& $0.17 \pm 0.05$ & $1.26 \pm 0.08$ \\
rewrite & & &	& & 	 & \\
\end{tabular}}
\caption{Various metrics of content preservation with standard deviations calculated across three paraphrase datasets, datasets of rewrites and various randomized datasets. GYAFC Formal and Informal correspond to the content preservation scores for GYAFC data treated as paraphrases in a formal or informal mode respectively. GYAFC and Yelp! rewrite correspond to the score between an input and a human-written reformulation in a different style. GYAFC and Yelp! random stand for the scores calculated on samples of random pairs from the respective dataset.}
 \label{tab:ex1}
\end{table*}

\begin{table*}[t!]
\centering
\tiny{\begin{tabular}{lllllllll}
 Dataset &	ELMo L2	& ROUGE-1 &	ROUGE-2	& ROUGE-L &	BLEU &	Meteor &	BERT score\\
 \hline
Bibles & $3.71 \pm 1.18$	& $0.61 \pm 0.17 $	& $0.38 \pm 0.22$ & $0.58 \pm 0.19$	& $0.28 \pm 0.24$ &	$0.6 \pm 0.2$ &	$0.93 \pm 0.03$ \\
Paralex & $5.74 \pm 1.41$ & $0.58 \pm 0.17$ & $0.24 \pm 0.2$ & 	$0.52 \pm 0.18$	& 	$0.07 \pm 0.17$ & 	$0.49 \pm 0.22$ & 	$0.91 \pm 0.03$ \\
Paraphrase & $6.79 \pm 1.87$ & $0.43 \pm 0.24$ & $0.13 \pm 0.22$	& $0.41 \pm 0.24$ & $0.01\pm 0.09$ 	& $0.4 \pm 0.27$ \ & $0.91 \pm 0.05$ \\
GYAFC informal & $5.56 \pm 1.31$ & $0.45 \pm 0.2$ & $0.22 \pm 0.19$	& $0.39 \pm 0.19$ & $0.1\pm 0.17$ 	& $0.4\pm 0.21$ \ & $0.89\pm0.04$ \\
GYAFC formal & $4.17 \pm 1.49$ & $0.61 \pm 0.21$ & $0.4 \pm 0.26$	& $0.57 \pm 0.22$ & $0.27\pm 0.28$ & $0.6\pm 0.23$ \ & $0.93\pm 0.04$ \\
\hline
Yelp! rewrite & $4.89 \pm 1.8$ & $0.57 \pm 0.24 $& $0.37 \pm 0.27	$&	$0.54 \pm 0.26$ & $0.22	\pm 0.28$ & $0.54 \pm 0.28$ & $0.92 \pm 0.04$ \\
GYAFC rewrite & $4.57 \pm 1.54$ & $0.61 \pm 0.21$ & $0.39 \pm 0.25$	& $0.56 \pm 0.22$ & $0.25 \pm 0.27$ 	& $0.57\pm 0.23$ \ & $0.92\pm0.04$ \\
\hline
Bibles random & $6.89 \pm 0.95$ & $0.15 \pm 0.07$ & $0.02 \pm 0.02$	& $0.12 \pm 0.06$ & $0.0\pm 0.0$ 	& $0.11\pm 0.06$ \ & $0.82 \pm 0.02$\\

Paralex random & $8.19 \pm 1.06$ & $0.23 \pm 0.11$ & $0.02 \pm 0.01$	& $0.21 \pm 0.11$ & $0.0\pm 0.0$ 	& $0.13\pm 0.1$ \ & $0.85 \pm 0.02$\\

Paraphrase random & $10.52 \pm 1.39$ & $0.03 \pm 0.01$ & $0.0 \pm 0.0$	& $0.02 \pm 0.01$ & $0.0\pm 0.0$ 	& $0.02\pm 0.01$ \ & $0.83 \pm 0.03$\\

GYAFC random & $7.67 \pm 1.02$ & $0.08 \pm 0.08$ & $0.01 \pm 0.03$	& $0.06 \pm 0.07$ & $0.01\pm 0.01$ 	& $0.06\pm 0.07$ \ & $0.82\pm 0.02$ \\
informal & & &	& & 	& & \\
GYAFC random & $7.55 \pm 0.92$ & $0.08 \pm 0.08$ & $0.01 \pm 0.03$	& $0.07 \pm 0.07$ & $0.000\pm 0.01$ 	& $0.08\pm 0.06$ \ & $0.84\pm0.02$ \\
formal & & &	& & 	& & \\
GYAFC random & $7.68 \pm 1.03$ & $0.07 \pm 0.07$ & $0.01 \pm 0.02$	& $0.06 \pm 0.06$ & $0.000\pm 0.000$ 	& $0.06\pm 0.05$ \ & $0.83\pm0.02$ \\
rewrite & & &	& & 	& & \\
Yelp! random& $8.19 \pm 0.9$ & $0.08 \pm 0.09$ & $0.002 \pm 0.02$ & $0.07 \pm 0.08$ & $0.000 \pm 0.000$	& $0.06 \pm 0.06$ & $0.85 \pm 0.02$ \\
rewrite & & &	& & 	& & \\
\end{tabular}}
\caption{Various metrics of content preservation  with standard deviations calculated across three paraphrase datasets, datasets of rewrites and various randomized datasets. GYAFC Formal and Informal correspond to the content preservation scores for GYAFC data treated as paraphrases in a formal or informal mode respectively. GYAFC and Yelp! rewrite correspond to the score between an input and a human-written reformulation in a different style. GYAFC and Yelp! random stand for the scores calculated on samples of random pairs from the respective dataset.}
 \label{tab:ex2}
\end{table*}

\section{Discussion}

Let us briefly discuss the desired properties of a hypothetical ideal content preservation metric. We do understand that this metric can be noisy and differ from dataset to dataset. However, there are two basic principles with which such metrics should comply. First, every content preservation metric that is aligned with actual ground truth semantic similarity should induce similar order on any given set of datasets. Indeed, let us regard two metrics $M_1$ and $M_2$ both of which claim to measure semantic preservation in two given parallel datasets $D_a$ and $D_b$. Let us assume that $M_1$ is the gold-standard metric that perfectly measures semantic similarity. Let us then assume that under the order that $M_1$ induces on the set of the datasets the following holds
$$M_1(D_a) \leq M_1(D_b).$$

Then either $$M_2(D_a) \leq M_2(D_b)$$ would be true in terms of the order induced by $M_2$ as well or $M_2$ is an inferior semantic similarity metric. 

Since style is a vague notion it is hard to intuitively predict what would be the relative ranking of style transfer pairs of sentences $D_s$, and paraphrase pairs $D_p$. However, it seems more than natural to disqualify any metric that induces such an order under which a randomized dataset ends up above the paraphrase or style transfer dataset. Under order induced by an ideal semantic preservation metric one expects to see both these datasets to be ranked above the dataset $D_r$ that consists of random pairs
\begin{equation}
\label{eq:crit}
M(D_r) \leq M(D_s);~~~ M(D_r) \leq M(D_p).
\end{equation}

Table \ref{tab:ex1} and Table \ref{tab:ex2} show resulting values of every metric across every dataset with standard deviations of the obtained scores. 

Table \ref{tab:order} summarizes order induced on the set of the paraphrase datasets, style transfer datasets, and datasets consisting of random pairs of sentences. One can see that humans rank random pairs as less semantically similar than paraphrases or style-transfer rewrites. Generally, human ranking corresponds to the intuition described in Inequalities \ref{eq:crit}. Majority of the metrics under examination are also in agreement with Inequalities \ref{eq:crit}.

What is particularly interesting is that humans assess GYAFC reformulations (the sentences with supposedly similar semantic but varying level of politeness) as the most semantically similar sentence pairs. However Yelp! rewrites that contain the same review of a restaurant but with a different sentiment are ranked as the least similar texts out of all non-random sentence pairs. This illustrates the argument made in \cite{TYwrong} that sentiment is perceived as an aspect of semantics rather than style by human assessors. Therefore, addressing the sentiment transfer problem as an example of the style transfer problem can cause systemic errors in terms of semantic similarity assessment. Unfortunately this often happens in modern style transfer research and should be corrected.

\begin{table*}[t!]
\centering
\tiny{\begin{tabular}{lllllllllllllll}
Metric & Bibles & Paralex & Paraphrase & Yelp! & GYAFC & GYAFC & GYAFC & Yelp! & GYAFC & GYAFC & GYAFC & Bibles & Paralex & Paraphrase\\

&random & random & random &random & random & random & random & rewrite & rewrite & informal & formal &&&\\

& &  &  &rewrite & rewrite &informal & formal & & & & & & &\\
 \hline
 POS & 14 & 10 & 8 & 9 & 11 & 12 & 13 & 1 & 4 & 7 & 2 & 5 & 6 & 3 \\
 Word overlap & 10 & 9 & 14 & 11 & 12 & 13 & 8 & 4 & 3 & 6 & 1 & 2 & 5 & 7 \\
 chrF & 9 & 10 & 14 & 11 & 12 & 13 & 8 & 4 & 2 & 7 & 3 & 1 & 5 & 6 \\
 Word2Vec & 8 & 12 & 14 & 11 & 7 & 10 & 9 & 4 & 2 & 5 & 3 & 1 & 6 & 13 \\
 FastText & 7 & 12 & 14 & 11 & 9 & 10 & 8 & 4 & 3 & 6 & 2 & 1 & 5 & 13 \\
 WMD & 8 & 13 & 14 & 11 & 10 & 9 & 12 & 4 & 1 & 6 & 3 & 2 & 5 & 7 \\
ELMo L2 & 8 & 13 & 14 & 12 & 11 & 10 & 9 & 4 & 3 & 5 & 2 & 1 & 6 & 7 \\
 ROUGE-1 & 10 & 9 & 14 & 11 & 13 & 12 & 8 & 5 & 3 & 6 & 1 & 2 & 4 & 7 \\
 ROUGE-2 & 10 & 9 & 14 & 13 & 12 & 8 & 11 & 4 & 2 & 6 & 1 & 3 & 5 & 7 \\
 ROUGE-L & 9 & 10 & 14 & 11 & 13 & 12 & 8 & 4 & 3 & 7 & 2 & 1 & 5 & 6 \\
 BLEU & 10 & 11 & 14 & 12 & 13 & 8 & 9 & 4 & 3 & 5 & 2 & 1 & 6 & 7 \\
 Meteor& 10 & 9 & 14 & 11 & 12 & 13 & 8 & 4 & 3 & 7 & 2 & 1 & 5 & 6 \\
 BERT score& 10 & 9 & 14 & 8 & 12 & 13 & 11 & 3 & 4 & 7 & 1 & 2 & 5 & 6 \\
 Human Labeling & 9 & 14 & 13 & 8 & 12 & 10 & 11 & 7 & 1 & 5 & 2 & 4 & 6 & 3 \\
\end{tabular}}
\caption{Different semantic similarity metrics sort the paraphrase datasets differently. Cosine similarity calculated with Word2Vec or FastText embeddings do not comply with Inequality $M(D_r) < M(D_p)$. All other metrics clearly distinguish randomized texts from style transfers and paraphrases and are in line with Inequalities \ref{eq:crit}. However, none of the metrics is completely in line with human evaluation.}
 \label{tab:order}
\end{table*}

\begin{figure*}[h!]
\centering
     \includegraphics[scale=0.45]{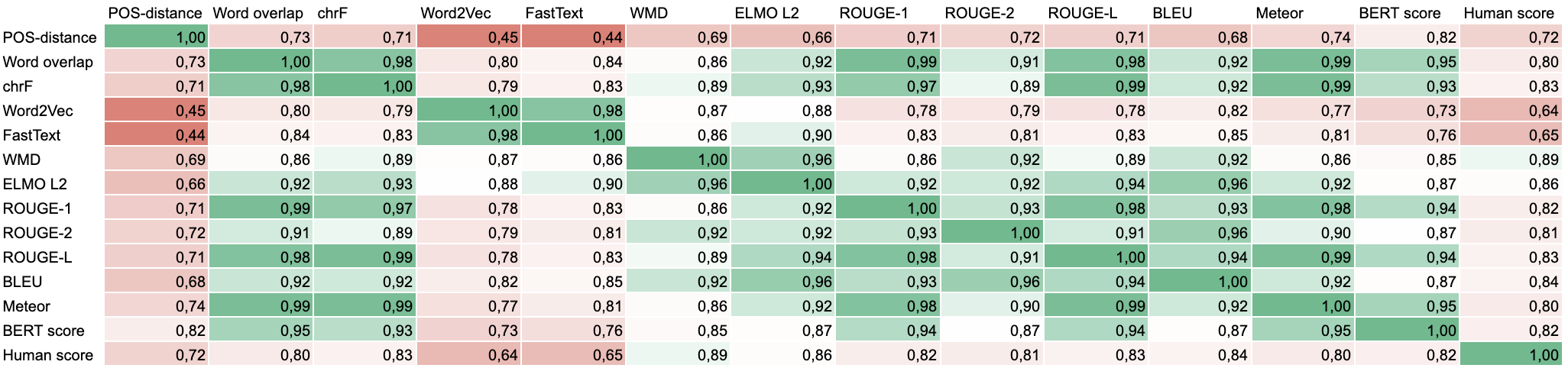}
  \caption{Pairwise correlations of the orders induced by the metrics of semantic similarity.}
  \label{fig:cor}
\end{figure*}

Closely examining Table \ref{tab:order} one can make several conclusions. First of all, cosine similarity metrics based on Word2Vec or on FastText do not seem to be useful as metrics of semantic preservation since they do not satisfy Inequality \ref{eq:crit} and also have the lowest correlation with human assesment, shown in Table \ref{tab:corr}. All the other metrics induce relatively similar orders on the set of the datasets. Figure \ref{fig:cor} illustrates that. 

Table \ref{tab:corr} shows correlation of the metric values with human assessments as well as correlations between human-induced order and the orders that other semantic similarity metrics induce. Table \ref{tab:corr} also demonstrates variability of the semantic similarity metrics. 

The intuition behind variability is to show how prone is the metric to fluctuations across different texts. Since on the datasets of random pairs the metric ideally should show very low semantic similarity, it is suboptimal if it assumes a large range of values on this datasets. The ratio between the range of values on random datasets and the range of values on all datasets is always between 0 and 1 plus and could intuitively characterize how noisy the metric is. If $\mathcal{R}$ is a set of all datasets of random pairs and $\mathcal{A}$ is set of all datasets in question, one can introduce a measure of metrics variability $V$ as

$$V = \frac{\max_{r \in \mathcal{R}}{M(D_r)}- \min_{r \in \mathcal{R}}{M(D_r)}}{\max_{a \in \mathcal{A}}{M(D_i)}- \min_{a \in \mathcal{A}}{M(D_i)}}.$$

For human labelling variability is relatively high $V = 19.7\%$ which means that humans often vary in their assessment of sentences that have no common semantic component. Lower variability on random pairs could be beneficial if one is interested in some form of binary classification that would distinguish pairs of sentences that have some information in common and the ones that do not. In this context BLEU seems to be superior to all other metrics of the survey. However, if we want to have some quantitative estimation of semantic similarity that resembles human judgement, than Meteor, chrF, and WMD seem to be more preferable.

\begin{table}[t!]
\centering
\scriptsize{\begin{tabular}{llll}
Metric 	& Correlation 			& Correlation & Variability\\
		& of the metric 	& of the induced & of the metric \\
		& with human  & orders with  & on random\\
		& evaluation & human-induced order & sentences\\
 \hline
 POS   &0.87 &  0.72 & 37.0\% \\
 Word overlap & 0.89 & 0.80 & 23.8\%\\
 chrF &  0.9 & 0.83 & 17.2\%\\
 Word2Vec  & 0.46 & 0.64 & 88.6\% \\
 FastText  & 0.52 & 0.65 & 86.3\%\\
 WMD   & {\bf 0.92} & {\bf 0.89} & 12.3\%\\
ELMo L2 & 0.82 & 0.86 & 53.3\%\\
 ROUGE-1  & 0.9 & 0.82& 33.5\%\\
 ROUGE-2 & 0.84 & 0.81 & 4.5\%\\
 ROUGE-L & 0.89 & 0.83 & 33.4\%\\
 BLEU & 0.72  & 0.84& 0.2\% \\
 Meteor & 0.91 & 0.80& 19.5\%\\
 BERT score & 0.89 & 0.82 & 23.1\% \\
\end{tabular}}
\caption{WMD shows the highest pairwise correlation with human assessment similarity scores.  The order on fourteen datasets, induced by WMD also has the highest correlation with human-induced semantic similarity order. Variability on random sentences is a ratio of the difference between the maximal and minimal value of a given metric on the datasets of random pairs and difference of the maximal and minimal value of the same metric on all available datasets.}
 \label{tab:corr}
\end{table}

One can also introduce several scoring systems to estimate how well every metric performs in terms of Inequalities \ref{eq:crit}. For example, we can calculate, how many datasets get the same rank in the metric-induced order as in the human-induced one. Another possible score could be a number of swaps needed to produce the human-induced order out of the metric-induced one. Table \ref{tab:scor} shows these scores for the the semantic similarity metrics in question. 

\begin{table}[t!]
\centering
\scriptsize{\begin{tabular}{lll}
Metric 	& Number of ranks		& Number of swaps\\
		& coinciding with		& needed to reconstruct\\
		& human-induced ranking & human-induced ranking\\
 \hline
 POS   &  3 &  16 \\
 Word overlap & 1 & 15\\
 chrF & 2 &  14 \\
 Word2Vec  & 3 & 16 \\
 FastText  & 2 & 17 \\
 WMD   & 1 & {\bf 11} \\
ELMo L2 & {\bf 4} & {\bf 11} \\
 ROUGE-1  & 0 & 15 \\
 ROUGE-2 & 2 & 13 \\
 ROUGE-L & 2 & 14 \\
 BLEU & 3 & 13\\
 Meteor & 2 & 15 \\
 BERT score & 3 & 13 \\
\end{tabular}}
\caption{Scores for the orders induced by different semantic similarity metrics.}
 \label{tab:scor}
\end{table}

Looking at the results listed above we can recommend the following. First of all, one has to conclude that there is no "silver bullet" for semantic similarity yet. Every metric that is used for semantic similarity assessment at the moment fails to be in line with human understanding of semantic similarity. It is important to add here that in terms of standard deviation human assessment is far more concise than some of the metrics under study. Though human scores vary from dataset to dataset the variance of them is relatively small when compared to the mean on any given dataset. Second, judging by Table \ref{tab:corr} and Table \ref{tab:scor} there are two metrics that seem to be the most promising instruments for the task. These are: WMD that induces the order with minimal amount of swaps needed to achieve human-induced order, shows the highest correlation with human assessment values, and the highest correlation with human-induced order; and ELMO L2 distance that has the highest number of coinciding ranks and is as well only eleven swaps away from a human-induced order, it also has the second highest in correlation for the induced order with the human-induced one, yet is relatively inferior in terms of pairwise correlation with human assessment.

Finally, let us look at Figure \ref{fig:cor}. There is a clear correlation between all orders induced by the metrics listed in Table \ref{tab:corr}. This correlation of induced orders is not only a consistent result that shows that the majority of semantic preservation metrics are aligned to a certain extent. This correlation could also be regarded as a justification of an order theory inspired methodology that we propose here for comparative analysis of metrics.

Looking at Figure \ref{fig:cor} one should also mention that POS-distance, as well as Word2Vec and FastText cosine similarities seem to be less aligned with every other metric that was tested. One could also see that WMD and ELMO L2 induce very similar orders. Taking this into consideration and revisiting results in Table \ref{tab:corr} and Table \ref{tab:scor} we can conclude that if one has to choose one metric of semantic similarity for a task of paraphrase or style transfer, WMD is the preferable metric at the moment.

The observed correlation of the induced orders gives hope that there is a universal measure of semantic similarity for texts and that all these metrics proxy this potential metric to certain extent. However, it is clear that none of them could model human judgement. There are several reasons that account for that. One is the phenomenal recent success of the semantic extraction methods that are based on local rather than global context that made local information-based metrics dominate NLP in recent years. Humans clearly operate in a non-local semantic context yet even state of art models in NLP can not account for this. The fact that BERT score that theoretically could model inner non-local semantics still does not reproduce human semantic similarity estimations is a proof for that. Second reason is the absence of rigorous, universally accepted definition for the problem of style transfer. We hope further research of disentangled semantic representations would allow to define semantic information in NLP in a more rigorous way, especially in context of several recent attempts to come up with unified notion of semantic information, see for example \cite{kolchinsky2018semantic}.

\section{Conclusion}

In this paper, we examine more than a dozen metrics for semantic similarity in the context of NLP tasks of style transfer and paraphrase. We publish human assessment for semantic similarity of fourteen thousand short text pairs and hope that this dataset could facilitate further research of semantic similarity metrics. Using very general order theory reasoning and human assessment data, we demonstrate that Word2Vec and FastText cosine similarity based metrics should not be used in context of paraphrase and style transfer. We also show that the majority of the metrics that occur in style transfer literature induce similar order on the sets of data. This is not only to be expected but also justifies the proposed order-theory methodology. POS-distance, Word2Vec and FastText cosine similarities are somehow less aligned with this general semantic similarity order. WMD seems to be the best semantic similarity solution that could be used for style transfer problems as well as problems of paraphrase at the moment. There is still no metric that could distinguish paraphrases form style transfers definitively. This fact is essential in the context of future style transfer research. To put that problem in the context of paraphrase, such semantic similarity metric is direly needed. 


\bibliographystyle{aaai}
\bibliography{acl}

\end{document}